\documentclass[lettersize,journal]{IEEEtran}
\usepackage{amsmath,amsfonts}
\usepackage{algorithmic}
\usepackage{algorithm}
\usepackage{array}
\usepackage[caption=false,font=normalsize,labelfont=sf,textfont=sf]{subfig}
\usepackage{textcomp}
\usepackage{stfloats}
\usepackage[hyphens]{url}
\usepackage{verbatim}
\usepackage{graphicx}
\usepackage{cite}
\usepackage{booktabs}
\usepackage{hyperref}
\begin{document}

\title{Automatic Speech Recognition for Speech Assessment of Persian Preschool Children}

\author{Amirhossein Abaskohi, Fatemeh Mortazavi, Hadi Moradi \\
School of Electrical and Computer Engineering, College of Engineering \\ University of Tehran, Tehran, Iran
\\ \texttt{\{amir.abaskohi, mortazavi.fatemeh, moradih\}@ut.ac.ir}
}



\maketitle

\begin{abstract}
Preschool evaluation is crucial because it gives teachers and parents influential knowledge about children's growth and development. The COVID-19 pandemic has highlighted the necessity of online assessment for preschool children. One of the areas that should be tested is their ability to speak. Employing an Automatic Speech Recognition (ASR) system would not help since they are pre-trained on voices that differ from children's in terms of frequency and amplitude. Because most of these are pre-trained with data in a specific range of amplitude, their objectives do not make them ready for voices in different amplitudes. To overcome this issue, we added a new objective to the masking objective of the Wav2Vec 2.0 model called Random Frequency Pitch (RFP). In addition, we used our newly introduced dataset to fine-tune our model for Meaningless Words (MW) and Rapid Automatic Naming (RAN) tests. Using masking in concatenation with RFP outperforms the masking objective of Wav2Vec 2.0 by reaching a Word Error Rate (WER) of 1.35. Our new approach reaches a WER of 6.45 on the Persian section of the CommonVoice dataset. Furthermore, our novel methodology produces positive outcomes in zero- and few-shot scenarios.\footnote{Code is publicly available at \url{https://github.com/AmirAbaskohi/Automatic-Speech-recognition-for-Speech-Assessment-of-Persian-Preschool-Children}}
\end{abstract}


\section{Introduction}
\label{sec:intro}

Before starting school, it is critical to have a thorough knowledge of children's skills. To acquire this knowledge, we can use an assessment system whose components should test a separate area of the child's abilities. Due to the COVID-19 pandemic, the demand for an online assessment system for preschool children was felt. The speech evaluation is one of the most challenging aspects of these tests. Speech recognition models are still not as accurate as Natural Language Processing (NLP) and Computer Vision (CV) models. Working with children's voices makes this assessment much more difficult.

We will compare several models in our in this article and share the dataset we used to train or fine-tune for Persian. Furthermore, we present a novel pre-training objective for the Wav2Vec 2.0 model\cite{baevski2020wav2vec}, resulting in state-of-the-art Persian ASR results.

ASR for children is a complex topic with many applications. Although there is a substantial body of work comparing the acoustic and linguistic features of children and adults \cite{lee1997analysis, kent1976anatomical}, our knowledge of how these variations impact speech recognition performance is limited.

Adult speech recognition has evolved substantially in recent years, but successful recognition of children's speech has been improved trivially\cite{shivakumar2014improving, gerosa2009review}. Identifying children's speech is challenging as children's capacity to recognize speech sounds correctly varies while they are learning to talk\cite{russell2007challenges, hamalainen2014correlating}. Moreover, transitions are weighted more strongly by children than adults for the fricative contrast\cite{mayo2004adult}. Children's general speaking pace is slower than adults, and their speaking rate, vocal effort, and spontaneity of speech are more diverse\cite{potamianos1997automatic}. It has been demonstrated that training directly on children’s speech may minimize this disparity in performance on a digit identification test, albeit accuracy is still inferior to that of adults\cite{elenius2005adaptation}. The spectral and temporal oral variability in children's speech is a key obstacle in acoustic modeling children's speech. Because formant values are more variable in children than adults, there is more interference across phonemic classes, making the categorization task intrinsically more complex\cite{potamianos1997automatic}. Furthermore, children have a far more comprehensive range of values for most auditory features than adults.
Five-year-old children, for example, have formant values that are up to 50\% greater than male adults\cite{lee1997analysis}. ASR performance can be severely harmed by a broad acoustic parameter range combined with increased acoustic unpredictability. 

\begin{figure}
  \centering
  \includegraphics[width=8cm,height=7cm,keepaspectratio]{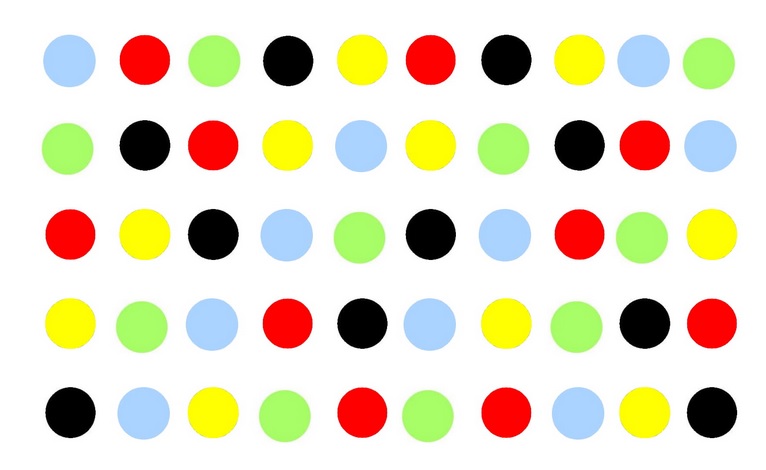}
  \caption{In the RAN test, the kid is given a series of objects, such as colors, and is asked to name them in order. The number of correctly detected objects determines the score of the test.}
  \label{fig:Rapid Automatic Naming Test}
\end{figure}

In our system, ASR is required for two tests:
\begin{itemize}

\item {The first test is called Rapid Automatic Naming (RAN). RAN is a behavioral test that evaluates how fast and accurately people name each group of visual stimuli. For fluent naming of a sequence of visual stimuli, RAN depends on the coordination of various processes into a synchronized access and retrieval mechanism (Figure \ref{fig:Rapid Automatic Naming Test})\cite{mcmillen2020rapid}. The difficulty in this activity includes word sequence speech classification and the necessity of correctness because the findings will be used to assess children's ability to communicate. In addition, due to the importance of speed in this activity, the model should provide accurate results rapidly. }

\item {The second test is a phonological memory test, so-called Meaningless Words (MW), in which youngsters are asked to listen to a meaningless word and then repeat it. This exam is complicated since it demonstrates strong developmental connections between test results and young children's vocabulary, reading, and overall abilities\cite{gathercole1994children}. The task's primary struggle is the remarkable similarity between these words and the actual words. For instance, the word "spoon" is called /qashoq/ in Persian. Using
/mashoq/ instead of /qashoq/ is an example of what happens in this task. These similarities make classification hard.}

\end{itemize} 

We propose a unique pre-training strategy for the Wav2Vec 2.0 model, called RFP, which adjusts the state-of-the-art Wav2Vec 2.0 model for various frequency domains in order to get a decent model for the tests mentioned above. The main contribution of this paper is twofold:

\begin{itemize}

\item {We propose a novel pre-training approach using pitch manipulation to create an ASR model for preschool children assessment in Persian.  }

\item {We show the benefits of the RFP objective in different domain frequencies and zero- and few-shot cases.  }

\end{itemize} 

\section{Related Work}
\label{sec:realtedwork}

Several strategies for improving voice recognition for children in English and other languages, including Persian in our instance, have been developed.

A number of studies have looked into the vocal differences between adults and children.
Mayo et al.\cite{mayo2004adult} showed that children's weigh transitions more heavily than adults for the fricative contrast. Also, they found that children weigh transitional cues more heavily than nontransitional cues for the voice-onset-time contrast.

The difference in vowel formants between children aged 7 to 9 and adults aged 18 to 22 was examined by Mohammadi et al.\cite{mohammadi2008persian} utilizing 25 girls and 25 boys for children and adults. Simple samples were used to create six Persian vowels (/i/,/e/,/a/,/o/,/u/,/â/). The first three formants of Persian language vowels were acquired and compared between male adults, school boys, and female adults and school girls. The findings revealed that disparities between children and adults are related to variations in vocal tract length and resonator cavity size.

ASR models must be modified for kids owing to the differences between adults' and children's voices. Ghai and Sinha et al.\cite{ghai2009exploring} implies that the conventional mel-spaced filter banks for generating recognition features are not up to the task. They noticed aberrations in the lower frequency filters in high-pitched speech and they suggested extending the filter bandwidth to smooth them out in the context of children's speech recognition. Mel-frequency Cepstral Coefficient (MFCC) features and a weaker Gaussian Mixture Model (GMM) were utilized in their trials.

Liao et al.\cite{liao2015large} investigated that speech recognition for children is not only a concern but also the expectation of an endless vocabulary system. Based on this concept, they developed a deployable Large Vocabulary Continuous Speech Recognition (LVCSR) system for children. They captured wide-band audio to account for the lengths of children's voice tracts so that high-frequency filter banks may be established. To design a child-friendly experience, they developed two properties for the language model: reduce the risk of objectionable mis-recognitions and better simulate the sorts of inquiries children were likely to utter. Ultimately, with models like LSTM and CLDNN\cite{sainath2015convolutional}, they achieved WER of 9.4\% and 20.0\% in clean and noisy training conditions.

Wav2Vec 2.0 is a cutting-edge ASR model. In the case of children's speech recognition, the earlier techniques use various methods to smooth the voices. Wav2Vec 2.0 can better learn the language from the speech signals since it employs rebuilding some masked parts of speech as a pre-training aim. Our approach is to keep this goal in mind while also including a new idea: learning the language regardless of voice pitch and frequency. With this idea, we are essentially attaining smoothness at the pre-training stage.

After the release of the Farsdat speech dataset in 1996\cite{bijankhan1994farsdat}, Persian ASR research became more serious. Research Center of Intelligent Signal Processing (RCISP) released the initial version of the Shenava ASR, built on a neural network, in 2001\cite{almasganj2001shenava}. This system had a 60\% accuracy rate for continuous voice recognition.

Based on the Hidden Markov Model (HMM), Sameti et al.\cite{sameti2008nevisa} has introduced Nevisa. In a normal environment, the suggested model's accuracy rate for continuous speech recognition was 75\%. In \cite{sameti2011large}, the second version of the Nevisa was released, achieving a 95\% accuracy rate for continuous speech recognition from independent speakers.

In 2016, On the Farsdat dataset, Daneshvar et al.\cite{daneshvar2016persian} employed DLSTM with a Connectionist Temporal Classification (CTC) output layer for Persian phoneme detection in 2016.

In 2020, Veisi et al.\cite{veisi2020persian} applied a mix of Deep Belief Networks (DBN) and Deep Bidirectional Long Short-Term Memory (DBLSTM) with CTC output layers for the acoustic model on the Farsdat dataset for the first time. The DBN employed in their model generated 39 unique properties for each frame. With 200 blocks and a learning rate of 0.0005, they achieved 16.7\% Phoneme Error Rate (PER). 

The Wav2Vec 2.0 model, which uses CTC layers that showed excellent performance in previous works and Transformer\cite{vaswani2017attention} architecture, displayed excellent results across various languages due to the masking pre-training objective. Even though we think that since people's tones vary from country to country, our pre-training objective can assist the model in adapting to various tones, leading to better outcomes in many languages.

In the field of adapting the ASR models for children, different approaches have been invested in previous works. Chen et al.\cite{chen2020data} used different data augmentation methods, including pitch perturbation augmentation, for children's speech recognition. Also, Shahnawazuddin et al.\cite{shahnawazuddin2020domain} created a robust automatic speaker verification for children when the domain-specific data is limited by using speed and pitch perturbation methods. Although the idea of using data pitch perturbation is not new, task-specific objective, which is sometimes those augmentation goals, is our method's novel idea. In Wav2Vec 2.0, we saw that not only contrastive learning helps a lot but also masking objective plays a crucial role, and it should not be considered the same as masking augmentation\cite{park2019specaugment, jain2021spliceout}.

\section{Proposed Method}
\label{sec:methodology}

As mentioned previously, our system includes a variety of speech recognition jobs, each with its unique set of features. We tested many models to get satisfactory results for these assessments and discovered their shortcomings. We thus decided to fix this problem by incorporating a new pre-training goal into Wav2Vec 2.0 model to modify this model for various frequency domains. Additionally, we collected our dataset using Persian children's voice records to fine-tune our model for our assessments.

\subsection{Dataset}
\label{subsec:dataset}
Because our assessments contain specific words and should be utilized with youngsters, we need to fine-tune our model on our own dataset. Furthermore, since one of the assessments comprises meaningless words, providing the model with this data is crucial in classification models.

Data collection was conducted by asking some adults from social media and some students from an elementary school to participate in our experiment.

Table \ref{table: colors data} shows the number of data gathered of each color for RAN test. Since there are two Persian terms for black, the number of black samples is more. In addition, because color recognition is a RAN task, some samples for this task have been gathered. Table \ref{table: colors sequence data} depicts the number of samples that contains a sequence of colors. For the MW assessment, 12 voices have been gathered on average per word (there are 40 meaningless words in the RAN test.)

\begin{table}
  \caption{The number of data gathered for each color in Persian. There are more black samples since there are two terms in Persian for black.}
  \label{table: colors data}
  \begin{center}
  \begin{tabular}{ccc}
    \toprule
    Color & Persian Phonetic & Number of Samples\\
    \midrule
    Blue & /abī/ & 483 \\
    Red & /qirmiz/ & 482 \\
    Black & /siyâh/ and /meškī/ & 873 \\
    Green & /sabz/ & 472 \\
    Yellow & /zard/ & 488 \\
    \bottomrule
  \end{tabular}
  \end{center}
\end{table}

\begin{table}
  \caption{Colors sequence data for RAN task. This data has been gathered in two noisy and quiet environments.}
  \label{table: colors sequence data}
  \begin{center}
  \begin{tabular}{cc}
    \toprule
    Environment & Number of Samples\\
    \midrule
    Quiet & 90 \\
    Noisy & 24 \\
    \bottomrule
  \end{tabular}
  \end{center}
\end{table}

\subsection{Rapid Automatic Naming Task}
\label{subsec:rantask}
In this task, the youngster should name a sequence of objects in an image. Since the outcome is crucial in evaluating the kid's ability to quickly name aloud a series of familiar items, classifying the entire sequence is insufficient; each word must be analyzed.

\begin{figure}
  \centering
  \includegraphics[width=8.5cm,height=7cm,keepaspectratio]{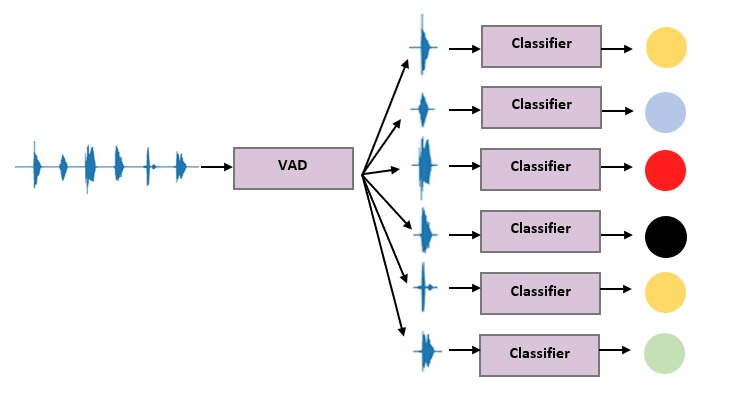}
  \caption{The architecture of the model we had used for RAN which contains VAD and CNN classifier. VAD detects each section of the voice in which a word is there, and then that word will be classified using the CNN model.}
  \label{fig:Combination of Varioational Encoder and a Classifier for RAN}
\end{figure}

We employed a mix of Vocal Activity Detection (VAD) and a Convolutional Neural Network (CNN) classifier as one of our strategies (Figure \ref{fig:Combination of Varioational Encoder and a Classifier for RAN}). 

We used MFCC features and a CNN model to classify each segment. Three convolutional layers with a dropout of 0.3, two dense, fully connected layers with a dropout of 0.3, and a softmax layer are utilized in this network. Even though this network achieved an accuracy of 92\%, there are issues with utilizing it to evaluate children. The main problem is the accuracy of VAD. 

\begin{figure}
  \centering
  \includegraphics[width=8.5cm,height=7cm,keepaspectratio]{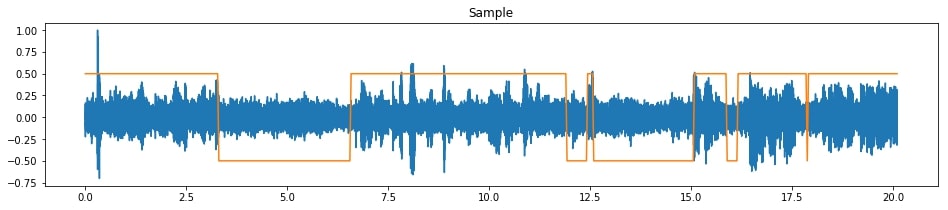}
  \caption{VAD is noise sensitive, so it may select an invalid word. For example, in 12.5s, a noise is selected as a word.}
  \label{fig:VAD sensitivity to noise}
\end{figure}

Noise enormously impacts VAD (Figure \ref{fig:VAD sensitivity to noise}). Moreover, the classifier can recognize which color was said, but determining whether the word is a color or not, is extremely difficult and it requires a large amount of data. Consequently, we decided to combine an ASR (which will be discussed further) with a text evaluation method to assess the youngster better.

\subsection{Meaningless Words Task}
\label{subsec:meaninglesstask}
In this task, the children must repeat a meaningless word that is played for them. These meaningless words are produced by modifying some parts of a word's phonemes to make them sound similar to the real word and unavailable in the Persian language's lexicon. For example /sacaroni/ instead of /macaroni/. This phoneme altering can be found anywhere in the word.

This task is different from the previous one. There is no sequence here, and a simple classifier can handle the output. We trained a CNN classifier similar to the model used in Section \ref{subsec:rantask} and attained a 90\% accuracy rate. The model was powerful enough to identify the word, however, it was not powerful enough to determine whether the word was valid. It was challenging to determine whether a word was invalid since we had to construct a new class for such terms. Many data samples should be gathered for this portion, particularly for words similar to those in our classes.

Since accuracy was crucial in all circumstances and the classification methods while having positive results, had several issues to be used in our system, we decided to test ASR models.

\subsection{Automatic Speech Recognition Approach}
\label{subsec:asrapproach}

We observed in the sections \ref{subsec:rantask} and \ref{subsec:meaninglesstask} that classifiers can not assist us as we require great accuracy in the desired models, and each test has its unique attributes. ASR can help us to reach our goal as they transform voice into text. However, this ASR model should handle few-shot situations. 

In the field of NLP, Transformers have shown excellent results. Wav2Vec 2.0 demonstrates how learning meaningful representations from voice audio alone and fine-tuning on transcribed speech surpasses the best semi-supervised approaches while being conceptually simpler with Transformer architecture.

Due to a self-supervised training method, a relatively novel idea in deep learning, Wav2Vec 2.0 becomes one of the most advanced models for ASR. With this training method, we may pre-train a model using unlabeled data, which is always easier to collect. The model may then be adjusted for a particular purpose on a given dataset.

Wav2Vec 2.0's pre-training goal is to mask input speech in the latent space and solve a contrastive task specified over a quantization of the joint learned latent representations. This objective is what happens in the T5 model's Masked Language Modeling (MLM)\cite{raffel2020exploring} objective.

The masking improves the model's performance in few-shot scenarios. However, our findings imply that in situations like ours, the model should be fine-tuned not only in a different language but also in the voices of youngsters. Even though Wav2Vec 2.0 performs well in this area, it is insufficient for our testing approach, which places a high weight on the model's performance. We proposed a new target for this model in Section \ref{subsec:RFP} as a result of this deficiency, allowing it to perform better in few-shot scenarios and various frequency domain circumstances.

\subsection{Pre-training With Random Frequency Pitch}
\label{subsec:RFP}

\begin{figure*}
    \centering
    \subfloat[\centering Original sound]{{\includegraphics[width=8cm]{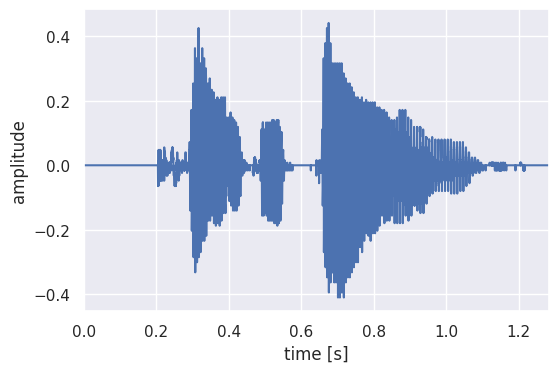} }}
    \qquad
    \subfloat[\centering Sound after using RFP algorithm]{{\includegraphics[width=8cm]{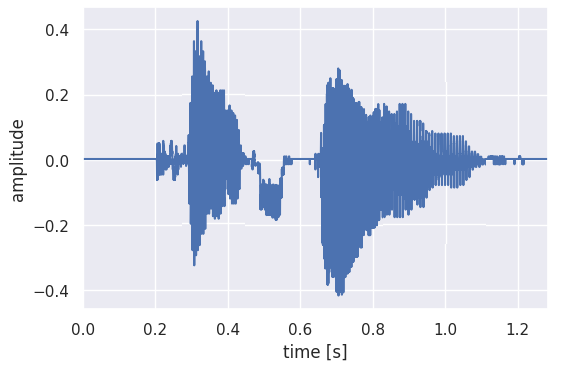} }}
    \caption{This is an example of using RFP on the voice. Chunks are created per 0.1s. The pitch manipulation changes the amplitude of the sound which results in Wav2Vec 2.0 frequency adaptation. Some 0.1s chunks of the original sound in (a) are selected and manipulated which resulted in (b).}
    \label{fig:RFPExample}
\end{figure*}

Wav2Vec 2.0 does not perform well enough in different frequency domains, which results in insufficient accuracy in our children's speech assessment\cite{sriram2022wav2vec}. We believe this poor performance is due to the pre-training approach of the Wav2Vec 2.0 model. 

A self-supervised learning approach is used in Wav2Vec 2.0. This method learns broad data representation from unlabeled instances before fine-tuning it with labeled data. Wav2Vec 2.0 is a framework for learning representations from raw audio data without supervision. This model uses a multi-layer CNN to encode spoken sounds, then it masks spans of the resultant latent speech representations.

This masking goal allows ASR to learn the language from voice signals and achieve state-of-the-art results. Although masking aids the learning of the language and performance in few-shot situations, it is unprepared for varied frequency domains. As a result, the model performs ineffectively when utilizing the model on children's voices. Thus, RFP, our pre-training method, aims to perform well in various frequency domains that could help us achieve better outcomes in these cases.

\begin{figure}
    \centering
    \centering
    \includegraphics[width=10cm,height=8cm,keepaspectratio]{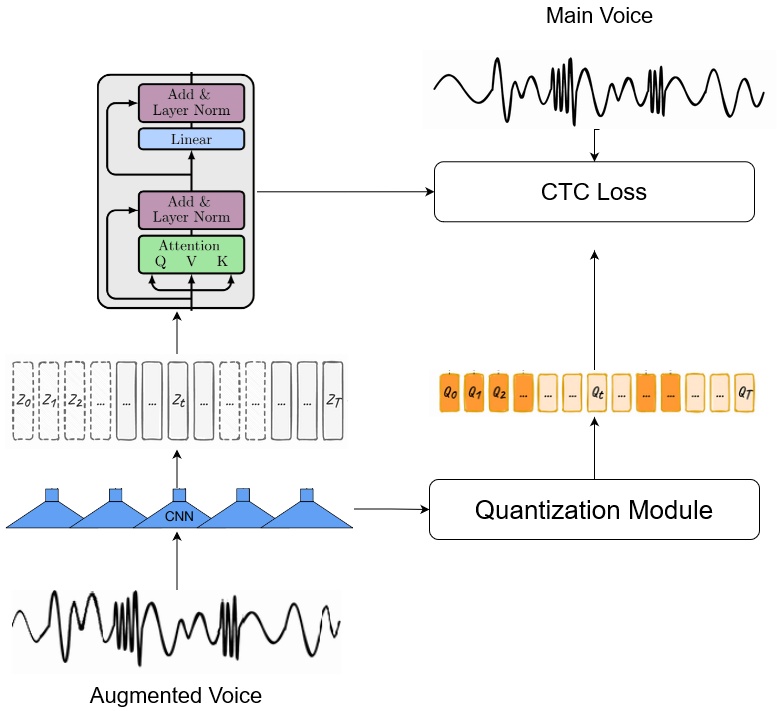}
    \caption{An overview of our model. Wav2Vec 2.0 model masks some parts of the speech and then uses CTC loss to find the masked part. In our approach, we create an augmented version of each speech sample in the dataset using RFP. However, instead of feeding the model with this new sample, we pass the augmented sample as the input of the model, and the model masks this sample and then tries to create the main not augmented sample, which is passed as output.}
    \label{fig:Model}
\end{figure}

Figure \ref{fig:Model} shows our approach to training the model. In this approach, for a given sample $X$, we first create an augmentation of that using an RFP algorithm called $X'$. Then it is passed to the model, and using CNN, a latent encoder, we reach some new features(${z_1, z_2, ..., z_T}$). Now some of these features are masked based on the masking approach of Wav2Vec 2.0. After that, these features are passed to the transformers' encoder. In the following step, the model tries to predict the masked features using CTC loss and having the encoded features and the result of the quantization module. Here, some features that are not masked are created based on the parts modified by the RFP algorithm. As a result, masked parts are predicted based on augmented and not augmented parts together. By this approach, the model is adapted for children's voices. 

RFP begins by dividing the speech file into one-second segments. Then it chooses a random number between 0 and 1 from a uniform distribution for each piece. If the produced random number is greater than the defined threshold (0.7 in our tests), it would use Praat\footnote{\href{https://www.fon.hum.uva.nl/praat}{https://www.fon.hum.uva.nl/praat}} commands to manipulate pitch using Parselmoutch python module \footnote{\href{https://parselmouth.readthedocs.io}{https://parselmouth.readthedocs.io}}. First, it uses the "To Manipulation" command with a time step of 0.01s, a minimum pitch of 75, and a maximum pitch of 600 for pitch manipulation. Then it extracts the pitch tier using the "Extract pitch tier" command. Finally, the output chunk is built using the retrieved pitch tier and a random factor between 0.1 and 4 obtained from a uniform distribution. The altered sound is the concatenation of the chunks (modified and unmodified). Algorithm \ref{alg:RFP} shows steps of RFP.

\begin{algorithm}
    \caption{Random Frequency Pitch}
    \label{alg:RFP}
    \begin{algorithmic}
        \STATE sound = readSoundFile(fileName)
        \STATE chunks = splitSoundFile(sound)
        \FOR {chunk in chunks}
            \STATE chanceOfPitchManipulation = uniformRandom(0, 1)
            \IF {chanceOfPitchManipulation $<$ 0.7}
                \STATE chunk = pitchManipulation(chunk)
            \ENDIF
        \ENDFOR
        \STATE result = concatChunks(chunks)
        \STATE result.saveSoundFile(fileName)
    \end{algorithmic}
\end{algorithm}

As a result of this algorithm, the main speech remains the same, except there are some frequency and amplitude changes in some parts of the voice(Figure \ref{fig:RFPExample}).

\section{Results}
\label{sec:typestyle}

We utilized the CommonVoice\cite{ardila2019common} and LibriSpeech\cite{panayotov2015LibriSpeech} datasets to assess our pre-training goal. We used the CommonVoice dataset in conjunction with our dataset, introduced in Section \ref{subsec:dataset} to fine-tune our model for our evaluation system. The results of each experiment will be discussed in the following sections.

\subsection{Pre-Training Comparison On English}
\label{subsec: pre-trained}

\begin{table*}
  \caption{Results of fine-tuning Wav2Vec 2.0 model with masking and RFP pre-training objectives on LibriSpeech test-clean, CommonVoice, and our dataset for RAN and MW test. As accuracy for classification was necessary  for tests, we reported the accuracy of the model for RAN and MW tests as well}
  \label{table: fine-tune result}
  \begin{center}
  \begin{tabular}{ccccl}
    \toprule
    Pre-training Objective & Dataset & WER & Classification Accuracy\\
    \midrule
    RFP & CommonVoice Persian & 6.458672 & - \\
    Masking & CommonVoice Persian & 8.451789 & - \\
    \midrule
    RFP & LibriSpeech test-clean & 1.356789 & - \\
    Masking & LibriSpeech test-clean & 1.794562 & - \\
    \midrule
    RFP & CommonVoice Persian + RAN's Samples & 4.561247 & 0.985749 \\
    Masking & CommonVoice Persian + RAN's Samples & 5.152465 & 0.876786 \\
    RFP & CommonVoice Persian + Meaningless Words' Samples & 4.124865 & 0.991245 \\
    Masking & CommonVoice Persian + Meaningless Words' Samples & 7.158496 & 0.842563 \\
    \bottomrule
  \end{tabular}
  \end{center}
\end{table*}

We trained the Wav2Vec 2.0 model with the masking technique and once with RFP in conjunction with the masking technique in English to test our innovative strategy in pre-training the model.

\begin{figure}
  \centering
  \includegraphics[width=8.5cm,height=7cm,keepaspectratio]{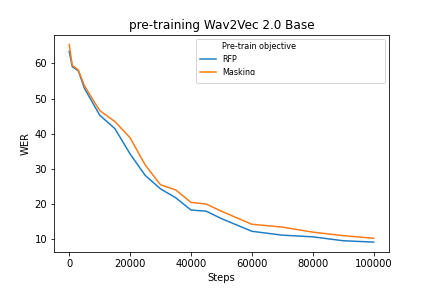}
  \caption{Results of pre-training Wav2Vec2.0-Base on LibriSpeech-960 dataset
for 100k steps with RFP objective and its comparison with pre-training with masking objective. The results are reported with WER. It can be seen that the RFP
objective outperforms the masking and converges faster.}
  \label{fig:pretrain}
\end{figure}

To mask the succeeding $M = 10$ time steps, we sample $p = 0.065$ of all time steps as beginning indices. As a result, almost 49\% of all time steps are masked, with an average span length of 14.7 or 299ms. The feature encoder is divided into seven blocks, each with 512 channels, strides of (5,2,2,2,2,2), and kernel widths of (10,3,3,3,3,2,2). This yields a 49hz encoder output frequency, a 20ms stride between samples, and a receptive field of 400 input samples or 25 milliseconds of audio. We use Adam\cite{kingma2014adam} to optimize, warming up the learning rate for the first 8\% of updates to a peak of $5 \times 10^{-4}$, and then linearly decaying it. Our model trains for 100k steps. For the diversity loss, we use the weight $\alpha = 0.1$. We utilize $G = 2$ and $V = 320$ for both models in the quantization module, resulting in a theoretical maximum of 102.4k codewords. The entries are $d/G = 128$ in size. Every update, the Gumbel softmax\cite{jang2016categorical} temperature $\tau$ is annealed by a factor of 0.999995 from 2 to a minimum of 0.5. In the contrastive loss\cite{baevski2020wav2vec}, the temperature is set to $\kappa = 0.1$. Batches are built by clipping 250k audio samples or 15.6sec, per example, using a base model with 12 Transformer blocks model dimension 768, inner dimension (FFN) 3072, and 8 attention heads. The models were trained using 960 hours of audio from the LibriSpeech corpus without transcriptions (LS-960). This experiment was carried out using Google Colab TPUs(v2-8). On LibriSpeech-960, our model not only converges quicker but also achieves superior WER (Figure \ref{fig:pretrain}).

\subsection{Supervised Speech Recognition For Assessments}
\label{subsec: fine-tuned}
First, to test our model on the stated evaluations, we fine-tuned the Wav2Vec 2.0 model, which is pre-trained on English LibriSpeech-960 with an RFP goal on Persian sections of CommonVoice.

In English, the model is evaluated on LibriSpeech-test-clean which is the test data of LibriSpeech. For Persian, we used 80\% of the data for training and the other 20\% for evaluation.

For comparison, we also fine-tuned the Wav2Vec 2.0 model, which was pre-trained with a masking aim. Then we used our dataset to fine-tune models for RAN and MW tests. In this experiment, Our collected samples were used to create more combinations of the samples. We created 100000 samples for the RAN test by selecting ten samples from each color and concatenating them with each other randomly. 10\% of this data and the samples mentioned in \ref{table: colors sequence data} were used as the test set. The remaining 90\% were used for fine-tuning the model and are shown in Table 1. All of the CommonVoice dataset has been used for fine-tuning. We used 380 samples for fine-tuning and the other 100 samples for testing in the MW test.

The model was fine-tuned for 50K steps with a batch size of 64, the same optimizer and learning rate as pre-training. The results are reported in Table \ref{table: fine-tune result}.

The results show that our model performs better on LibriSpeech as it can learn both language and vocal domains. In addition, due to the masking objective in the Wav2Vec 2.0 model, it can perform well after fine-tuning on other languages.

\subsection{Zero-shot and Few-Shot Speech Recognition Results}
\label{subsec: zero-few}


\begin{table}
  \caption{The zero- and few-shot speech recognition results of the models were evaluated with WER on the Persian section of CommonVoice.}
  \label{table: zero result}
  \begin{center}
  \begin{tabular}{ccc}
    \toprule
    Fine-tuning Steps & RFP & Masking\\
    \midrule
    0 & 37.865974 & 42.299586 \\
    10k & 30.484689 & 33.756942 \\
    20k & 26.652178 & 31.188419 \\
    30k & 13.127486 & 15.121548 \\
    40k & 11.298463 & 14.689485 \\
    50k & 10.418498 & 12.498441 \\
    \midrule
    zero-shot & 30.484689 & 33.756942 \\
    \bottomrule
  \end{tabular}
  \end{center}
\end{table}

We tested the pre-trained models in a zero-shot scenario to see how well they perform in resource-limited circumstances.
Without fine-tuning, each pre-trained model is assessed on a test set of the Persian part of CommonVoice.
The findings in Table \ref{table: zero result} reveal that our RFP target yields better outcomes in zero-shot speech recognition than the masking pre-training objective due to distinct domain frequency adaptation.

For few-shot experiments, we fine-tuned our model on 15h samples from the Persian section of the CommonVoice dataset at a maximum of 50K steps. The results in Table \ref{table: zero result} show that RFP outperforms masking in zero-shot scenarios, because of the simultaneous use of RFP and masking, which makes the model adaptable to any new environment.

\section{Conclusion}
\label{sec:conclusion}

In this study, we initially looked at the difficulties of children's voice recognition. Then we discussed our preschool cognitive tests in speech criteria and many options for achieving good model performance in these critical examinations. Despite their high accuracy, we concluded that categorization models cannot assist us in these assessments. So utilized Wav2Vec 2.0, a state-of-the-art ASR model, and we discovered that it does not perform well in different frequency domains. Consequently, we created a new pre-training goal for this model called RFP, which employs frequency pitching to modify the voice. Then, we discovered that the new model works well for children's voices and exceeds the masking objective. Our new objective also reaches better results in zero- and few-shot scenarios.
Future research should consider the potential effects of other pre-training approaches used in domains like NLP. For example, the effect of masking with reordering can be examined in this model.

\bibliographystyle{IEEEtran}
\bibliography{refs} 

\vfill

\end{document}